\documentclass[letterpaper, 10 pt, conference]{ieeeconf}  % Comment this line out if you need a4paper

\IEEEoverridecommandlockouts                              % This command is only needed if 
\usepackage{amsmath,graphicx}
\usepackage[utf8]{inputenc}
\usepackage{here,multirow,setspace,url}
\usepackage{fancybox}
\usepackage{amsmath,amssymb, algpseudocode}
\usepackage{enumerate,psfrag}
\usepackage{bm}
\usepackage{times,type1cm, pifont}
\usepackage{psfrag}
\usepackage{cite}
\usepackage{lscape}
\usepackage{color,subcaption, textcomp}
\usepackage{multirow}
\usepackage{algorithm} 
\usepackage{comment, booktabs, colortbl}
\usepackage[table]{xcolor}
\usepackage{rotating}

\title{\LARGE \bf
Whom to Respond To? A Transformer-Based Model for Multi-Party Social Robot Interaction
}

\author{He Zhu$^{1, 2}$, Ryo Miyoshi$^{1}$ and Yuki Okafuji$^{1}$% <-this % stops a space
\thanks{\newline$^{1}$ AI Lab, CyberAgent Inc., Tokyo 150-0042, Japan\newline $^{2}$ Hokkaido University, Sapporo, Japan}
}

\begin{document}

\maketitle
\thispagestyle{empty}
\pagestyle{empty}

%%%%%%%%%%%%%%%%%%%%%%%%%%%%%%%%%%%%%%%%%%%%%%%%%%%%%%%%%%%%%%%%%%%%%%%%%%%%%%%%
\begin{abstract}
Prior human-robot interaction (HRI) research has primarily focused on single-user interactions, where robots do not need to consider the timing or recipient of their responses. However, in multi-party interactions, such as at malls and hospitals, social robots must understand the context and decide both when and to whom they should respond. In this paper, we propose a Transformer-based multi-task learning framework to  improve the decision-making process of social robots, particularly in multi-user environments. Considering the characteristics of HRI, we propose two novel loss functions: one that enforces constraints on active speakers to improve scene modeling, and another that guides response selection towards utterances specifically directed at the robot. Additionally, we construct a novel multi-party HRI dataset that captures real-world complexities, such as gaze misalignment. Experimental results demonstrate that our model achieves state-of-the-art performance in respond decisions, outperforming existing heuristic-based and single-task approaches. Our findings contribute to the development of socially intelligent social robots capable of engaging in natural and context-aware multi-party interactions.
\end{abstract}
%%%%%%%%%%%%%%%%%%%%%%%%%%%%%%%%%%%%%%%%%%%%%%%%%%%%%%%%%%%%%%%%%%%%%%%%%%%%%%%%
\section{INTRODUCTION}
Human-robot interaction (HRI)~\cite{breazeal2004social} is becoming increasingly important as social robots integrate into daily life. Beyond enhancing task efficiency, social robots must engage in natural and context-aware communication to address societal challenges, such as aging populations and labor shortages. As societies across the globe face demographic shifts and evolving workforce dynamics, innovative technological solutions are essential to support sustainable human-robot collaboration~\cite{yamazaki2012home}.
Significant progress has been made in enabling social robots to understand and respond to human speech and emotions in one-on-one interactions~\cite{suzuki2022augmented}. 
However, through real-world experiments, we observed that multi-party HRI (MHRI) dominates in environments such as malls~\cite{koike2025drivesinteractroleuser, song2024new}. In one-on-one settings, interactions can often be managed through straightforward turn-taking mechanisms. In contrast, multi-party interactions require a deeper understanding of the scene to determine response strategies~\cite{skantze2021turn, gu2022says}. This includes recognizing overlapping speech, identifying intended addressees, and understanding contextual cues such as gaze direction and body orientation. Without these capabilities, interaction can easily break down, limiting the robot’s effectiveness in dynamic social environments.
Despite this, there is a critical gap in MHRI research. Addressing this gap is essential for developing the next generation of socially intelligent social robots~\cite{moujahid2022multi}.

A key challenge in MHRI is decision-making on whether the robot should respond from user utterance and, if so, to whom. In real-world multi-party interactions, users may switch between addressing the robot and other humans. For example, in a shopping mall, one person might ask the robot for directions while another, standing nearby, casually comments about the store layout. If the robot responds to every detected speech, it risks disrupting human-to-human interactions or misidentifying the intended addressee. Thus, effective response decision-making is crucial for ensuring meaningful engagement and maintaining conversational coherence. 
Much of the existing work~\cite{wei2023multi} often relies on simple heuristics, such as if-then rules~\cite{paul2023enhancing}, or employs large language models (LLMs) without deeply integrating multimodal cues~\cite{addlesee2024multi}. 
%Furthermore, many decision-making strategies depend heavily on gaze for response selection. However, real-world interactions frequently exhibit mismatches between gaze direction and conversational target, complicating accurate inference.
%
Recognizing the scene is an essential component of this decision-making process. Graph-based methods~\cite{ortega2018graph} and BERT-based speaker identification models~\cite{chen2021neural} have been used to model human-to-human social dynamics. However, as these methods do not fully address the challenges of HRI, where robots must infer intent with limited information and adapt dynamically to evolving conversations.
Beyond methodological challenges, the scarcity of large-scale datasets that simulate natural multi-party interactions further hinders progress. Data collection and annotation for such scenarios remain labor-intensive, limiting the ability to develop and evaluate robust, context-aware decision-making models for robots.

To address these challenges, we leverage GPT’s ability to model long-range dependencies and generate contextually relevant responses. While GPT-based~\cite{radford2019language} approaches have shown effectiveness in structured dialogues and sequential decision tasks, most existing works focus on single-user settings or text-based interactions, limiting their adaptability to complex, multimodal HRI. 
%Incorporating multimodal cues into the decision-making process can improve response selection and interaction coherence in dynamic conversations.
%
Multi-task learning enables the model to learn shared representations across tasks, improving feature extraction and reducing overfitting compared to training each task independently. In the context of MHRI, this allows the model to refine conversational role recognition while simultaneously optimizing response selection. Therefore, to further enhance decision-making, we adopt a multi-task learning framework that jointly trains scene recognition and response selection. By enforcing consistency between perception and decision-making, our approach ensures more context-aware and socially appropriate robot responses.

In this study, we addressed the novel challenge of response decision-making in MHRI. Unlike traditional HRI settings that focus on single-user interactions, our work explores how a social robot can determine whether to respond and to whom in multi-party scenarios. To address this challenge, we propose a multi-task learning framework based on GPT-2. Our approach first identifies the speaker and listener using encoded video and speech-to-text inputs, then leverages this information to guide the robot’s response decisions. To enhance learning, we introduce two loss functions: one enforcing single-speaker constraints for realistic scene modeling and another prioritizing responses directed at the robot for more appropriate engagement.
Additionally, we collected a novel dataset containing MHRI scenarios, featuring diverse interaction modes and contextual cues. Specifically, we designed scenarios where gaze direction aligns with or diverges from the speaker's intended target, simulating real-world complexities in multi-party interactions. Our proposed method was tested on this dataset and compared against state-of-the-art approaches. Experimental results demonstrated the effectiveness of our framework in both scene recognition and response decision tasks, highlighting its potential for advancing MHRI. 
Finally, we summarize our contributions of this paper as follows:
\begin{itemize} 
\item We introduce the novel problem of response decision-making in MHRI and propose a multi-task learning framework, incorporating two loss functions to enhance speaker recognition and prioritize appropriate robot responses.
\item We construct a new dataset simulating diverse multi-party interaction scenarios, incorporating real-world complexities, and use it to evaluate our proposed approach.
\item We demonstrate the effectiveness of our approach through comprehensive experiments, achieving state-of-the-art performance in scene recognition and response decision tasks. 
\end{itemize}
\section{RELATED WORK}
%Our research divides the robot's response decision into two parts: scene recognition and the response itself.
\subsection{Multi-Party Interaction Speaker-Listener Recognition}
Identifying speaker-listener relationships in multi-party interactions is essential for effective dialogue modeling. Gu et al.~\cite{gu2022says} conducted a survey on multi-party conversations, highlighting the challenges of speaker addressee recognition and turn-taking. Accurate addressee detection enables more natural and coherent interactions, particularly in dynamic social settings.

Graph-based approaches, such as ChatMDG~\cite{chatmdg2024} and JRDB-Social~\cite{jrdbsocial2024}, map conversational dynamics. Ortega et al.~\cite{ortega2018graph} further incorporated these cues into interaction models, demonstrating that multimodal information enhances recognition accuracy. However, these methods primarily focus on human-human interactions and do not account for the asymmetries inherent in HRI, where robots must infer conversational intent with limited contextual awareness. However, existing approaches often rely on structured conversational cues and predefined speaker roles, limiting their adaptability to dynamic, multi-party human-robot interactions where turn-taking and addressee recognition require real-time contextual inference.
Data collection plays a critical role in advancing multi-party conversational modeling. Li et al.~\cite{li2021thoughtful} emphasized the importance of structured corpora, noting that existing datasets lack sufficient diversity in multi-party settings. Li et al.~\cite{li2020interview} created a large-scale media dialogue corpus that captures complex turn-taking structures, offering insights into real-world speaker engagements. However, such datasets primarily focus on human interactions, and their direct applicability to HRI remains limited.
\subsection{Multi-Party Human-Robot Interaction}
MHRI requires robust decision-making strategies that go beyond single-user interactions. Early research explored engagement recognition and user classification~\cite{engagement2017} to assist in turn-taking, providing foundational insights into multi-party dialogue management. 

Recent advancements have leveraged LLMs for adaptive interaction management~\cite{wei2023multi, multipartychat2023}, enhancing the ability of robots to process context and adjust responses dynamically. However, existing LLM-based approaches are limited integration of crucial multimodal cues.
Robots deployed in social environments must dynamically adjust to conversational cues and determine appropriate intervention points. Studies on, e.g., multi-party interaction with a robot receptionist~\cite{hri2022} and multi-party medical consultations~\cite{eacl2024}, highlight domain-specific challenges, demonstrating the need for more sophisticated decision-making frameworks. However, many existing systems still rely on heuristic-based approaches~\cite{paul2023enhancing}, limiting adaptability in dynamic and unstructured environments.
Moreover, distinguishing intended addressees in multi-party interactions remains challenging due to misalignments between gaze direction and conversational targets. Some approaches attempt to infer speaker intent based on head pose and prosody, but these methods often fail in naturalistic, noisy environments. To Address this limitation, deeper integration of multimodal information and learning-based frameworks capable of handling uncertainty in human-robot communication are required.

Our work builds upon the abovementioned studies by introducing a multi-task learning framework that integrates multimodal cues with Transformer-based architectures. Focusing on response decision-making while incorporating scene recognition modeling as a supporting task through multi-task learning, our approach enhances the robot’s ability to make informed and context-aware responses in complex multi-party scenarios. This improves the effectiveness of MHRI, bridging the gap between theoretical advancements and real-world application.
%%%%%%%%%%%%%%%%%%%%%%%%%%%%%%%%%%%%%%%%%%%%%%%%%%%%%%%%%%%%%%%
\section{METHOD}
%%%%%%%%%%%%%%%%%%%%%%%%%%%%%%%%%%%%%%%%%%%%%%%%%%%%%%%%%%%%%%%%%
\subsection{Problem Formulation}
\begin{figure*}
    \centering
    \includegraphics[width=0.8\linewidth]{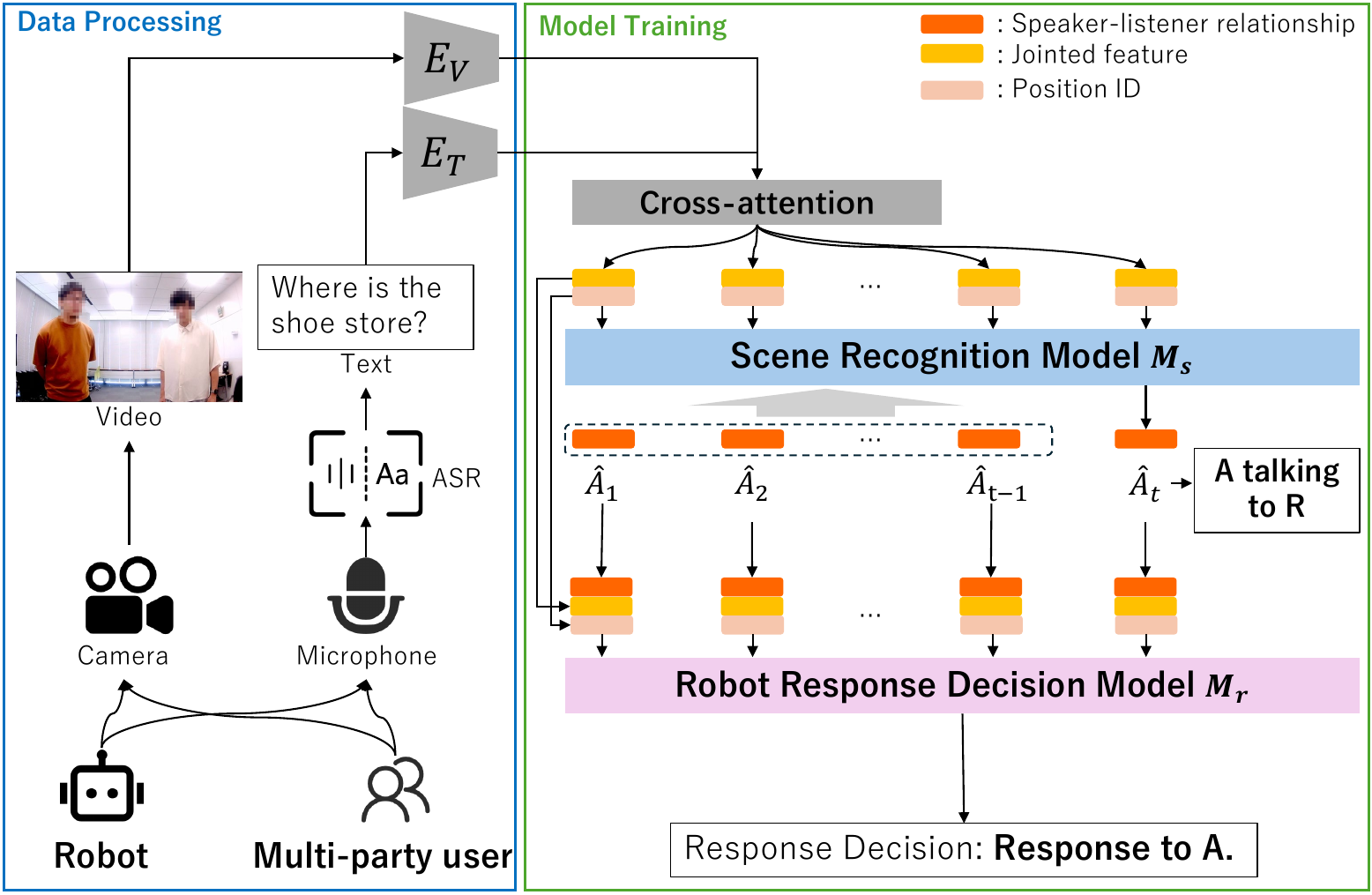}
    \caption{Overview of our MHRI framework. The system processes video and text input, extracts features, and uses a Transformer-based model to understand dialogue. The Robot Response Decision Model then determines whether and to whom the robot should respond.}
    \label{fig:overview}
\end{figure*}
We formalize MHRI as a structured sequence prediction problem.
We employ a multi-task learning framework, which consists of a scene recognition model $M_s$ and response decision-making model $M_r$.
Given a recorded MHRI video as input, $M_s$ must identify the active speaker and infer the intended listener, while $M_r$ must decide whether the robot should respond, and determine the appropriate target of its response. These models are jointly optimized, allowing the proposed method to leverage shared multimodal representations and improve prediction consistency across different conversational scenarios.

To address this new task, we collected a realistic scenario dataset consisting of a collection of MHRI videos $\{V_g\}_{g=1}^{G}$, where $G$ is the total number of data. A given video $V_g$ contains $N_g$ instances $I_n = \{S_n, A_n, R_n\}$ ($n$ = 1, ..., $N_g$), where $S_n$ is an extracted video segment, $A_n$ represents the speaker-listener relationship, and $R_n$ represents the response decision.
To formally describe our method, we define $P$ as the conditional probability distribution representing the likelihood of a response or interaction given the multimodal input. Specifically, the probability of a response decision can be computed as:
\begin{equation}
   P_r = P(R_n|S_n; \theta_d, \theta_r),
\end{equation}
where $\theta_s$ and $\theta_r$ are the parameters of $M_s$ and $M_r$, respectively. Given that HRI exhibits strong prior distributions due to structured conversational norms, we innovatively incorporate Kullback-Leibler (KL) divergence regularization to ensure that the model adheres to these prior patterns while maintaining adaptability.
Using this formulation, our method can be expressed as:
\begin{equation}
   \arg\min_{\theta_d, \theta_r} \sum_{g=1}^{G}\sum_{n=1}^{N_g} \mathbb{E}_{S_n \sim V_g} \Big[ \mathcal{L}(R_n, P_r) + 
   \lambda_s \mathcal{L}_{\text{KL}}^{\text{s}} + \lambda_r \mathcal{L}_{\text{KL}}^{\text{r}}\Big],
\end{equation}
where $\mathcal{L}_{\text{KL}}^{\text{s}}$ and $\mathcal{L}_{\text{KL}}^{\text{r}}$ represent the KL divergence with respect to their respective prior distributions. $\mathcal{L}$ is a joint objective function that integrates both scene recognition and response decision-making tasks. It captures the discrepancy between the predicted response decision \( R_n \) and the ground truth, incorporating multiple task-specific losses.

The proposed method integrates scene recognition and response decision-making in a unified optimization process. By structuring MHRI as a sequence prediction task, our approach effectively captures conversational dependencies and dynamically adapts robot responses to complex social interactions. This design enables the model to leverage multimodal context more efficiently, ensuring more accurate and contextually appropriate robot interventions in multi-user environments.
%%%%%%%%%%%%%%%%%%%%%%%%%%%%%%%%%%%%%%%%%%%%%%%%%%%%%%%%%%%%%%%%%%%
\subsection{Dataset Collection}
%%%%%%%%%%%%%%%%%%%%%%%%%%%%%%%%%%%%%%%%%%%%%%%%%%%%%%%%%%%%%%%%%%%
\subsubsection{Data Acquisition}
To investigate HRI, we collected a dataset featuring interactions between a humanoid robot (Sota) and two human participants. The data acquisition process was designed to capture natural multi-party interactions in a controlled setting, simulating public environments such as shopping malls and customer service areas. 
%%%%%%%%%%%%%%%%%%%%%%%%%%%%%%%%%%%%%%%%%%%%
\begin{figure}[t]
    \centering
    \includegraphics[width=0.6\linewidth]{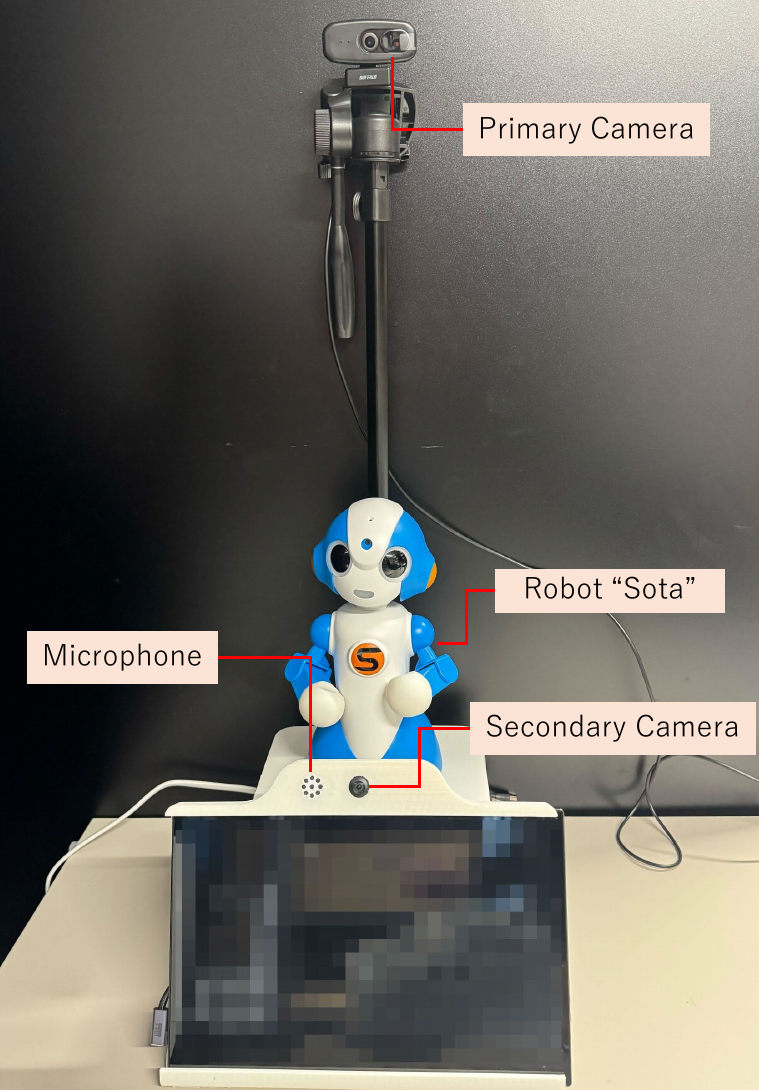}
    \caption{The experimental setup consists of the robot ``Sota" equipped with a primary camera, a secondary camera, and microphone.}
    \label{fig:robot}
\end{figure}
\begin{figure}[t]
  \centering
      \begin{minipage}[b]{0.55\linewidth}
            \includegraphics[width=\linewidth]{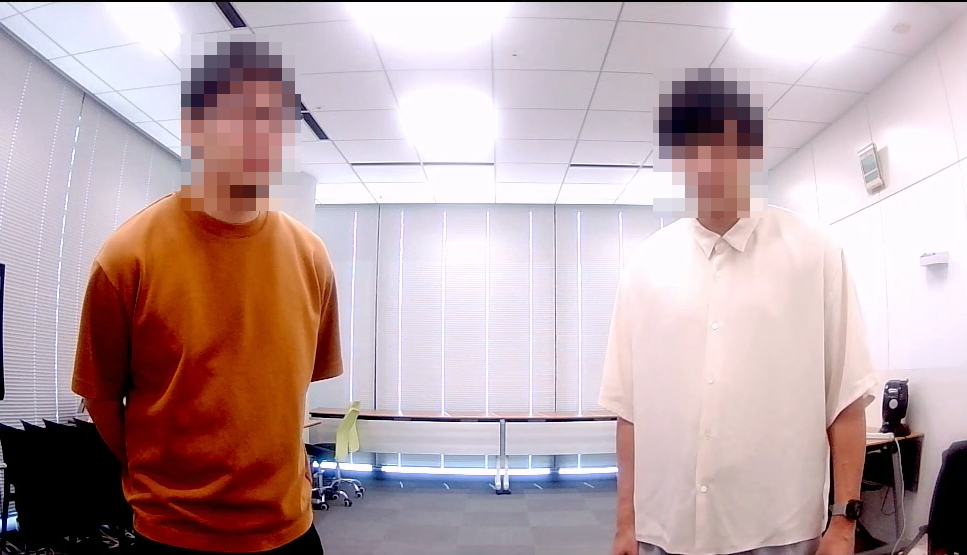}
            \subcaption{}
            \label{Fig:data sample 1}
    \end{minipage}
    \begin{minipage}[b]{0.42\linewidth}
            \includegraphics[width=\linewidth]{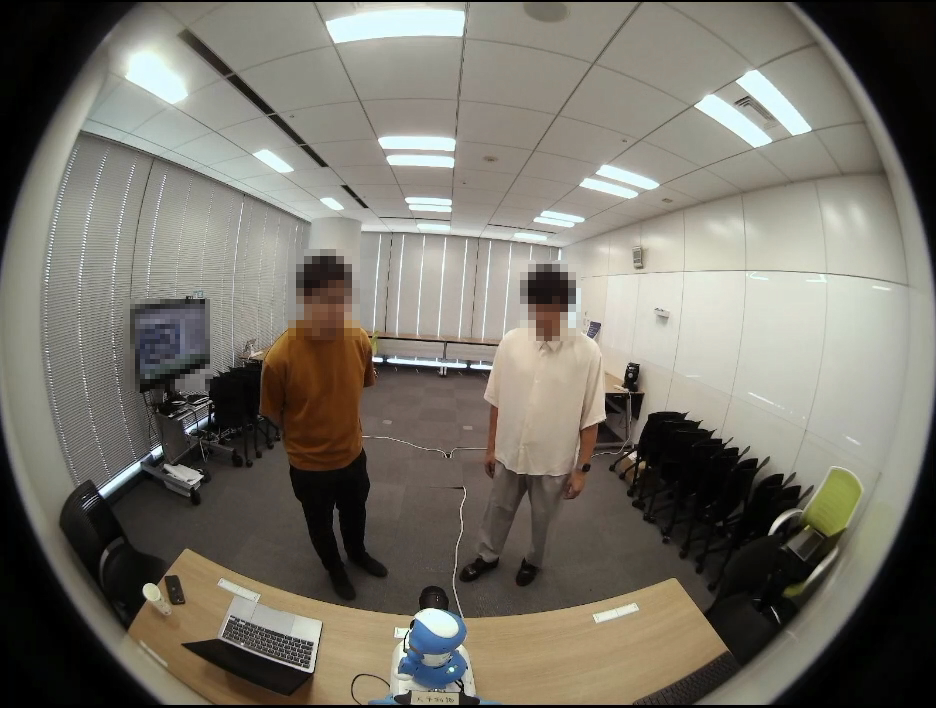}
            \subcaption{}
            \label{Fig:data sample 2}
    \end{minipage}
    \caption{An example of acquired video data. (a) Primary camera that captures the entire scene, and (b) secondary camera from the robot's perspective.} 
\end{figure}
\begin{figure}[t]
    \centering
    \includegraphics[width=\linewidth]{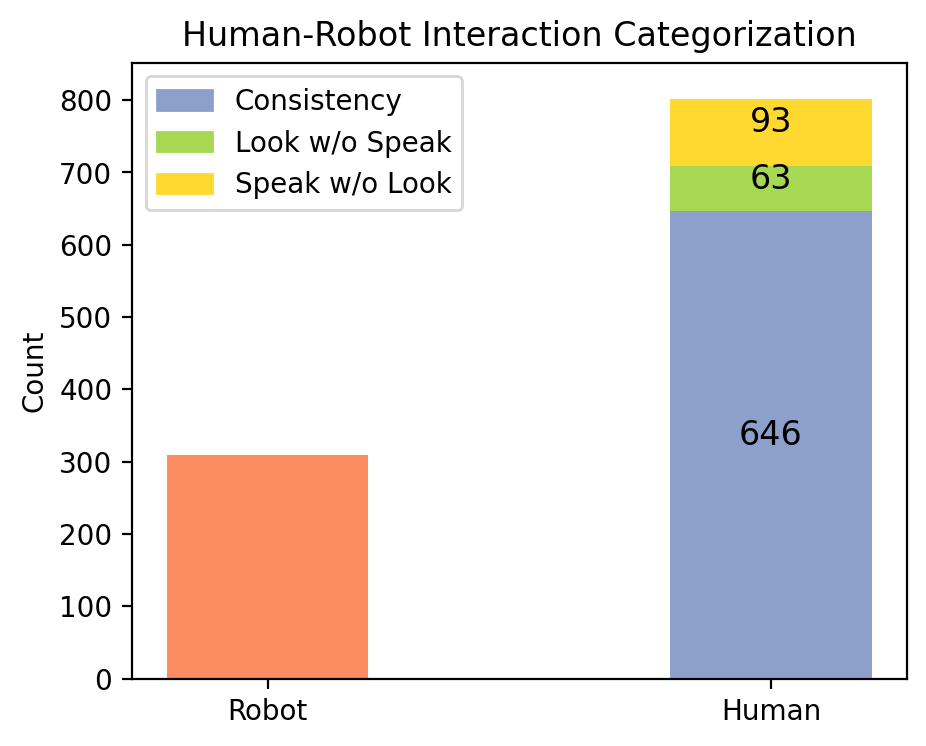}
    \caption{Distribution in our dataset. This bar chart represents the number of utterances in the dataset categorized by speaker type: Robot and Human. The stacked bars for human interactions are further divided into three categories: Consistency (the speaker looks at and speaks to the listener), Look w/o Speak (the speaker looks at the listener but does not speak), and Speak w/o Look (the speaker speaks to the listener without making eye contact).}
    \label{fig:enter-label}
\end{figure}
%%%%%%%%%%%%%%%%%%%%%%%%%%%%%%%%%%%%%%
\begin{table*}[t]
\centering
\caption{Example annotations from a single video in the dataset. ``A" and ``B" represent human users, while ``R" represents the robot. Each row corresponds to an utterance and includes the speaker, listener(s), Gaze Alignment, indicating the speaker's visual focus, and Response Decision, indicating whether the robot should respond.}
\begin{tabular}{c|c|c|c|c|c}
\toprule
\rowcolor[HTML]{E6E6E6} 
Idx & Content                               & Speaker & Linser(s)  & Gaze Alignment &  Respond Decision \\\midrule
1   & There's a guide robot.                & A       & B       & A looks at R  &  No   \\ 
2   & Wow, that's amazing.                  & B       & A       & B looks at R  &  Yes   \\
3   & Hello. Are you looking for something? & R       & A, B & R looks at A, B  &  No   \\
4   & Where is the shoe store?              & A       & R       & A looks at R  &  No   \\
... & ...                                   & ...     & ...     & ...               \\\bottomrule
\end{tabular}
\label{Table:data sample}
\end{table*}
%%%%%%%%%%%%%%%%%%%%%%%%%%%%%%%%%%%%%%%
As shown in Fig.~\ref{fig:robot}, the robot, which is approximately 100 cm in height, was remotely operated by a human to generate natural responses. Two cameras and a microphone were deployed to record multi-modal interaction data. 
As shown in Fig.~\ref{Fig:data sample 1} and \ref{Fig:data sample 2}, we used a dual-camera setup to capture interaction dynamics effectively. The primary camera was used for training purposes, providing a comprehensive scene representation to analyze speaker-listener relationships and interaction flow. The secondary camera, positioned at the robot’s eye level, approximated the robot’s visual perspective and was primarily intended to assist the operator in better recognizing the surrounding environment. Additionally, a microphone was placed near the robot to ensure clear speech capture for speech-to-text processing.

To ensure the dataset’s robustness and representativeness, we carefully designed and validated the interaction scenarios based on existing MHRI studies and preliminary pilot trials. Each session lasted approximately 3 min and followed two types of interaction scenarios: task-oriented interactions, where participants engaged with the robot to seek specific information, and casual interactions, where participants engaged in open-ended dialogues without predefined goals. There were casual interactions in 16.7\% of cases. As shown in Fig.~\ref{fig:enter-label}, we categorized utterances based on gaze alignment: ``Consistency", where the speaker's gaze is towards the listener, and unaligned gaze, which includes two cases—``Look w/o Speak", where the speaker looks at one listener but speaks to another, and ``Speak w/o Look", where the speaker speaks to a listener without making eye contact. Additionally, we considered 10\% cases where utterances were directed towards multiple listeners simultaneously, making it challenging to determine the primary recipient of the speech. 
\subsubsection{Data Annotation}
To enable supervised learning for both speaker-listener identification and robot response decision-making, we developed a two-stage annotation pipeline, consisting of automatic annotation followed by manual refinement. This hybrid approach significantly improved efficiency while ensuring high-quality labels.

The automatic annotation process leveraged Google Cloud Speech-to-Text with speaker diarization enabled to generate initial transcriptions and speaker segmentation. Given the limitations of automated speech recognition (ASR) systems in handling overlapping speech and segmentation errors, we incorporated an additional post-processing step using GPT-4o to refine the outputs. Specifically, the transcription pipeline consisted of the following steps:
\begin{enumerate}
    \item Extract audio from recorded video files and convert them into WAV format for standardized processing.
    \item Perform speech recognition using Google Cloud Speech-to-Text with multi-speaker diarization, enabling automatic segmentation of individual utterances.
    \item Process the transcription results with GPT-4o, refining sentence boundaries, correcting phonetic misrecognition, and aligning timestamps with utterance-level segmentation.
\end{enumerate}

Following the automatic transcription stage, human annotators performed manual verification and correction. Annotators also corrected instances where the ASR system failed to distinguish overlapping speech or misattributed utterances, and each utterance's start and end timestamps were manually verified and corrected to ensure precise alignment with the speech data. Additionally, to ensure annotation consistency and reliability, each utterance was labeled by multiple annotators, and in cases of disagreement, the final label was determined using a majority vote strategy. Furthermore, the dataset was enriched with ``robot response annotations", where annotators labeled whether the robot should respond. The labeled data are shown in the Table~\ref{Table:data sample}.

This annotation framework resulted in a structured multi-modal dataset containing synchronized video, audio, and transcriptions, along with speaker-listener interaction labels. The resulting dataset provides a valuable resource for training and evaluating models in multi-party HRI scenarios, particularly for tasks involving conversational dynamics understanding and response decision-making.
%%%%%%%%%%%%%%%%%%%%%%%%%%%%%%%%%%%%%%%%%%%%%%%%%
\subsection{Training Process}
%%%%%%%%%%%%%%%%%%%%%%%%%%%%%%%%%%%%%%%%%%%%%%%%%
The overview of the proposed method is shown in Fig.~\ref{fig:overview}. Given a recorded MHRI video, we extract multimodal features from both visual and textual data. Specifically, we utilize a Transformer-based encoder to process video frames, capturing spatial and motion dynamics, while an ASR system transcribes speech into text. The extracted text is then embedded using a pre-trained language model to obtain semantic representations. 

Let $\mathbf{V} \in \mathbb{R}^{T \times D_v}$ represent the video features, $\mathbf{T} \in \mathbb{R}^{T \times D_t}$ the text embeddings, where $T$ denotes the sequence length, and $D_v$ and $D_t$ are the respective feature dimensions.

\subsubsection{Scene Recognition Model}
The extracted features are processed by a scene recognition model, denoted as $M_s$, to identify the active speaker and their intended listener. We employ a Transformer decoder architecture with cross-attention mechanisms to integrate multimodal representations effectively. The model takes the multimodal features as input and predicts a structured sequence representing speaker-to-listener mappings.

The prediction of speaker-listener relationships is formulated as:
\begin{equation}
P(A_n | \mathbf{V}_n, \mathbf{T}_n; \theta_d) = M_s(\mathbf{V}_n, \mathbf{T}_n; \theta_s),
\end{equation}
where $A_n$ represents the speaker-listener relationship, and $\theta_s$ are the model parameters. 

To enhance learning stability, we introduce a structured loss function that enforces constraints on speaker activation. Specifically, we consider two complementary loss terms: a cross-entropy loss for speaker identification and a KL divergence-based regularization to regulate conversational flow. The cross-entropy loss ensures accurate speaker predictions:
\begin{equation} 
\mathcal{L}_{\text{CE}}^{\text{s}} = \sum_{n=1}^{N} \text{CrossEntropy}(A_n, \hat{A}_n). 
\end{equation}
Meanwhile, to reflect natural conversational patterns, we introduce a KL divergence loss that discourages a single speaker from speaking continuously without interruption. This regularization term encourages natural turn-taking dynamics:
\begin{equation} 
\mathcal{L}_{\text{KL}}^{\text{s}} = D_{\text{KL}}(P(\hat{A}_n) \parallel P_{\text{prior}}(A_n)), 
\end{equation}
where $P_{\text{prior}}(A_n)$ represents an empirically derived prior that discourages consecutive speaking turns by the same individual. The overall speaker modeling loss is then formulated as:
\begin{equation} 
\mathcal{L}_{\text{s}} = \mathcal{L}_{\text{CE}}^{\text{s}} + \lambda_{\text{s}} \mathcal{L}_{\text{KL}}^{\text{s}}, 
\end{equation}
where $\lambda_{\text{s}}$ is a hyperparameter that controls the relative weight of the KL regularization term.

By incorporating both loss terms, our model not only achieves accurate speaker prediction but also aligns conversational structures with natural human interactions. This structured speaker modeling serves as a foundation for the subsequent Robot Response Decision Model, where the predicted speaker-listener relationships are leveraged to determine whether and how the robot should respond.
\subsubsection{Robot Response Decision Model}
The predicted speaker-listener relationships are further processed by the response decision-making model, denoted as $M_r$. The response decision-making model operates on the speaker-listener relationships predicted by the previous module, leveraging the structured conversational context learned from $M_s$ alongside multimodal features to produce response probabilities:
\begin{equation}
P(R_n | \mathbf{V}_n, \mathbf{T}_n, \hat{A}_n; \theta_r) = M_r(\mathbf{V}_n, \mathbf{T}_n, \hat{A}_n; \theta_r),
\end{equation}
where $R_n$ represents the robot’s response decision.

To ensure that the robot's responses adhere to conversational norms, we incorporate two key loss components: a cross-entropy loss to optimize response prediction and a KL divergence-based regularization to enforce interaction constraints. The classification loss is defined as:
\begin{equation} \mathcal{L}_{\text{CE}}^{\text{r}} = \sum_{n=1}^{N} \text{CrossEntropy}(R_n, \hat{R}_n). \end{equation}
Beyond accuracy, conversational appropriateness is crucial. We introduce a KL divergence term that encourages the robot to prioritize responding to individuals directly addressing it while preventing extended self-initiated turns:
\begin{equation} \mathcal{L}_{\text{KL}}^{\text{r}} = D_{\text{KL}}(P(R_n | S_n, \theta_r) \parallel P_{\text{prior}}(R_n)), \end{equation}
which encodes empirical interaction patterns, guiding the robot to engage in a socially coherent manner. This ensures that responses align with conversational expectations rather than being arbitrary or excessively frequent. The final objective for response modeling is:
\begin{equation} \mathcal{L}_{\text{r}} = \mathcal{L}_{\text{CE}}^{\text{r}} + \lambda_{\text{r}} \mathcal{L}_{\text{KL}}^{\text{r}}, \end{equation}
which regulates the influence of regularization on response selection.

This module ensures that the robot’s responses are not only contextually relevant but also socially appropriate by explicitly incorporating structured conversational norms. Traditional approaches to robot response selection often rely on handcrafted if-then rules, where responses are determined by direct signals without explicitly modeling the conversational structure. Furthermore, the KL-based constraint introduces a novel mechanism to regulate response behavior, ensuring that the robot prioritizes direct interactions while avoiding excessive self-initiated turns. These structured constraints improve the naturalness and efficiency of multi-party human-robot interactions, making the system more adaptive to real-world social dynamics.
\subsubsection{Joint Optimization}
The primary goal of our framework is to enable the robot to make contextually appropriate and socially aware response decisions in multi-party interactions. To achieve this, we integrate structured dialogue modeling as a supporting step, ensuring that response selection is informed by the conversational context rather than relying solely on surface-level cues. To jointly optimize these components, we define a unified objective function:

\begin{equation} \mathcal{L} = \mathcal{L}_{s} + \mathcal{L}_{r} , \end{equation}

Traditional methods for robot response selection are often rule-based, relying on heuristics such as gaze tracking or simple turn-taking rules to determine responses. These approaches fail in ambiguous situations, where conversational cues are less explicit or involve multiple participants.
Our proposed framework overcomes these limitations by introducing a data-driven, structured approach to response selection. Instead of treating response selection as an isolated classification problem, we incorporate structured scene recognition to ensure that responses align with real-world conversational patterns. By introducing KL divergence-based constraints, we explicitly enforce that the robot prioritizes responding to individuals addressing it directly, reducing inappropriate or contextually irrelevant responses.
This avoids excessive self-initiated turns, maintaining natural conversational flow and preventing robotic interruptions.

By focusing on response decision-making as the core objective, our framework addresses the fundamental challenge of MHRI determining when the robot should engage and with whom—leading to more intelligent, natural, and socially aware interactions.
\section{EXPERIMENTS}
\begin{table*}[t]
\centering
\caption{Comparison of accuracy (AC) and per-decision time across different methods for determining the appropriate response target. ``G $\neq$ L" represents cases where the speaker's gaze and speaking target are different, while ``G = L" indicates alignment between gaze and speaking target, and ``Average" denotes the overall accuracy. }
\begin{tabular}{l|c|ccc|c}
\toprule
\rowcolor[HTML]{E6E6E6} 
Method & Type   & AC (G $\neq$ L)  & AC (G = L)  & AC (Averange) &  Per-Decision Time \\\midrule
Paul et al.~\cite{s23135798}           & If-then & 30.0   & 65.5    & 59.5  &  0.1s   \\ 
Addlesee et al.~\cite{addlesee2024multi}            & Modeling     & 25.0   & 71.7    & 63.8  &  5.4s   \\
GPT-4 (text only)      & Modeling     & 22.7   & 55.4    & 48.6  &  1.0s   \\
GPT-4o-mini   & Modeling     & 40.8   & 68.0    & 62.7  &  5.3s  \\
GPT-4o        & Modeling     & 22.4   & 67.0    & 58.2  &  5.9s   \\
GPT-o1-mini (text only)  & Modeling     & 28.6   & 58.0    & 52.2  &  9.2s   \\\midrule
PM            & Modeling& \textbf{60.0}   & \textbf{72.2}    & \textbf{66.2}  &  \textbf{0.01s}   \\\bottomrule
\end{tabular}
\label{Table:results}
\end{table*}
\subsection{Experimental Setup}
To evaluate our proposed method, we conducted experiments in an MHRI setting. The objective was to determine whether the robot should respond at all and who the appropriate recipient is. 
We constructed a dataset consisting of multi-party interactions in a mall-like environment, where human participants engage in both purposeful and casual interactions with the robot. Our dataset consists of 60 videos, with a total duration of 82 min and a total of 1,111 dialogues. Each interaction is labeled with the speaker’s identity and the intended recipient of the utterance. Our data includes both cases where the gaze direction is aligned and misaligned with the speaking target, with the specific distribution shown in Fig.~\ref{fig:enter-label}. Additionally, 10\% of the sentences involve speaking to multiple recipients simultaneously.

We used the pre-trained InternVideo2 model~\cite{wang2024internvideo2} to extract video and text features and employed cross-attention to fuse information from different modalities, which ensures robust multimodal alignment between video and text. This facilitates improved speaker identification and response prediction in multi-party human-robot interaction scenarios. The proposed method is based on a GPT-2 architecture that processes multimodal inputs, including video and textual features. The model was trained using the AdamW optimizer with a learning rate of 0.0001, a batch size of 8, a epoch of 30, and a dropout rate of 0.1. We adopted a 6-fold cross-validation strategy to ensure robustness. We set both hyperparameters $\lambda_{\text{s}}$ and $\lambda_{\text{r}}$ to 0.01 based on experimental results. Furthermore, we introduced a multi-ignore cross-entropy loss function that filters out non-relevant classes, such as padding tokens and robot utterances, to improve training efficiency. We used a Nvidia L4 GPU (24G memory) with Pytorch 2.5.0 version.

To assess the effectiveness of our approach, we compared it with two pervious methods: those by Addlesee et al.~\cite{addlesee2024multi} and Paul et al.~\cite{s23135798}. The former integrates a LLM with gaze information, while the latter relies solely on an if-then rule-based approach using gaze information. Additionally, we compared our method with different versions of ChatGPT, including those supporting multimodal input and those limited to text input. For the text-only versions, we incorporated dialogue context and scene information into the prompt. Our evaluation was primarily based on the accuracy of the response decision and the required time. For accuracy, we measured cases where the gaze direction aligns or misaligns with the speaking target separately, as well as the overall accuracy. To further investigate the contributions of key components, we conducted ablation studies to analyze the impact of different loss functions and the effectiveness of the multi-task learning framework.
\subsection{Experimental Results}
\begin{table}[t]
\centering
\caption{Ablation study results for the proposed model in MHRI. M denotes the use of multi-task learning, while $\mathcal{L}_{\text{KL}}^{\text{s}}$ and $\mathcal{L}_{\text{KL}}^{\text{r}}$ represent the two KL divergence losses incorporated in the training process.}
\begin{tabular}{c|ccc|ccc}
\toprule
\rowcolor[HTML]{E6E6E6}
&  &  & & \multicolumn{3}{c}{Accuary}\\ \cline{5-7} \rowcolor[HTML]{E6E6E6}
 \multirow{-2}{*}{} & \multirow{-2}{*}{M} & \multirow{-2}{*}{$\mathcal{L}_{\text{KL}}^{\text{s}}$} &  \multirow{-2}{*}{$\mathcal{L}_{\text{KL}}^{\text{r}}$} & (G $\neq$ L)  & (G = L)       & (Avg)         \\ \midrule
(a)&              &  &    & 60.3          & 65.5          & 64.5          \\
(b)&              &  & \ding{51}  & 55.1          & 66.9          & 64.6          \\ \midrule
(c)& \ding{51}    &  &      & 62.2          & 66.7          & 65.8          \\
(d)& \ding{51}    &       & \ding{51}       & 60.3          & 65.0          & 66.0          \\
(e)& \ding{51}    & \ding{51}    &          & \textbf{62.2} & 67.0          & 66.1          \\
(f)& \ding{51}     & \ding{51}   & \ding{51}  & 60.0        & \textbf{72.2} & \textbf{66.2} \\ \bottomrule
\end{tabular}
\label{Table:ablation}
\end{table}
\begin{figure}[t]
  \centering
    \begin{minipage}[b]{0.49\linewidth}
            \includegraphics[width=\linewidth]{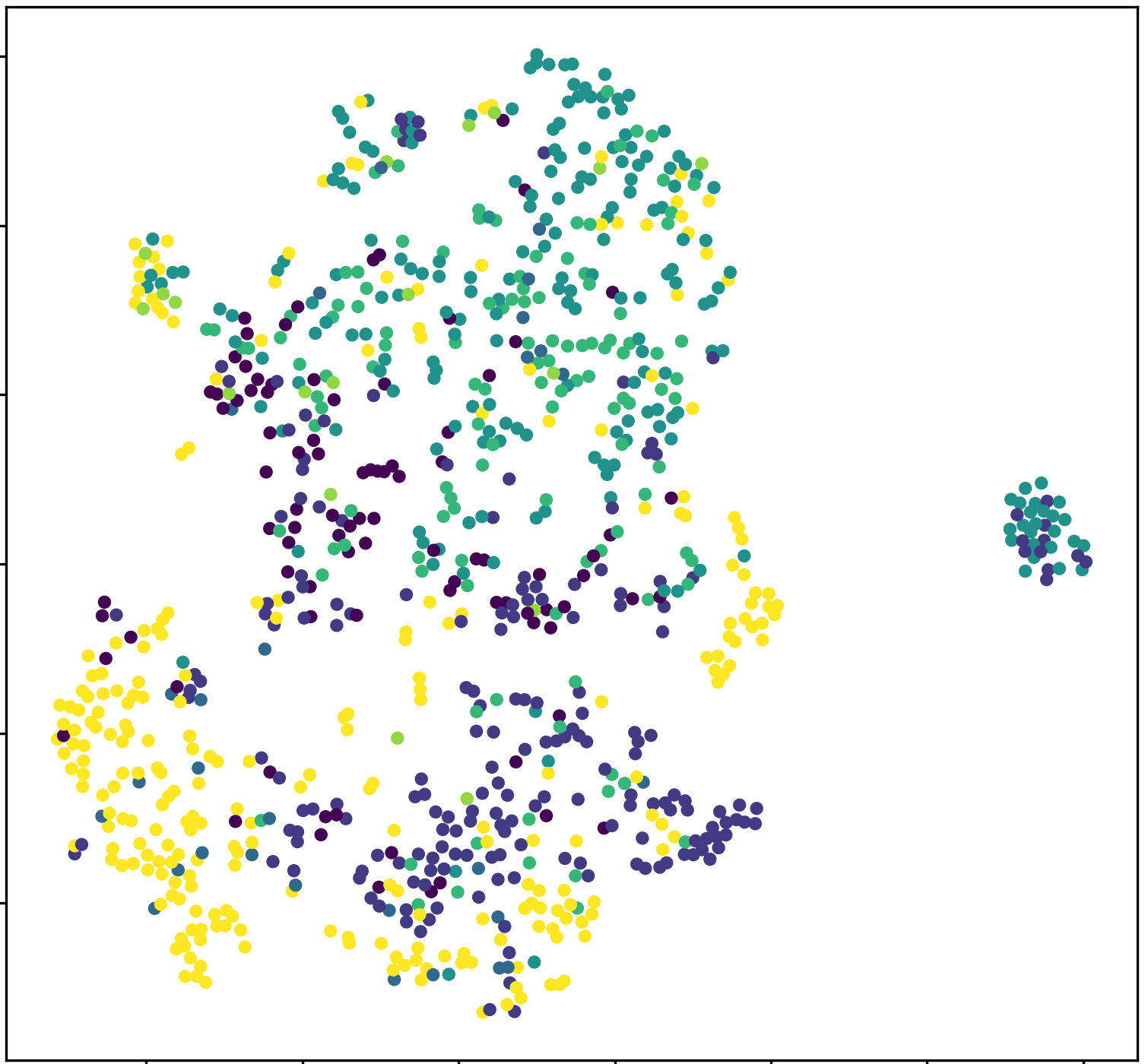}
            \subcaption{}
            \label{Fig:visualize a}
    \end{minipage}
    \begin{minipage}[b]{0.49\linewidth}
            \includegraphics[width=\linewidth]{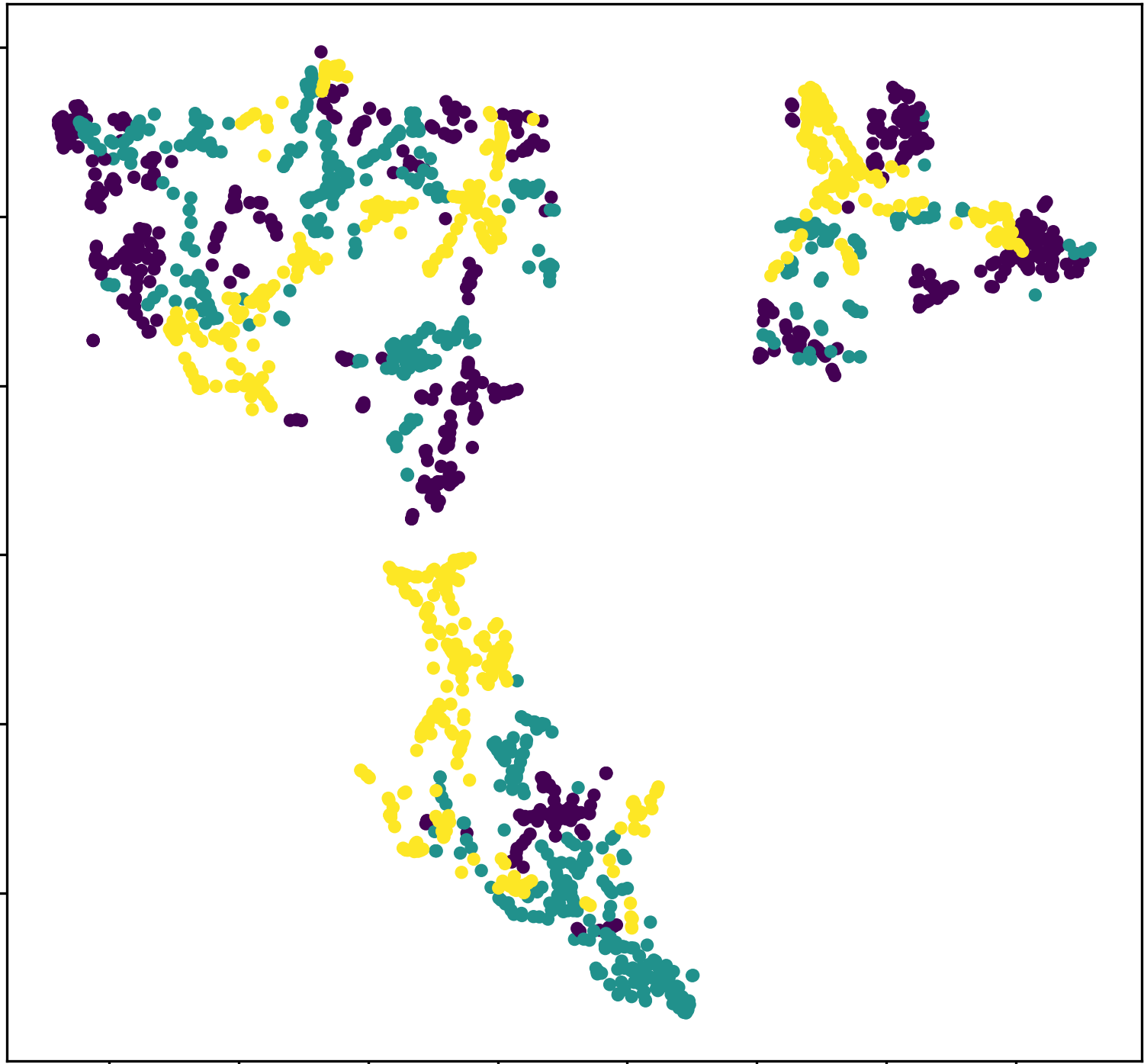}
            \subcaption{}
            \label{Fig:visualize b}
    \end{minipage}
    \caption{ t-SNE visualization of fused features for (a) scene recognition and (b) response decision. Different colors represent distinct categories.} \label{Fig:visualize}
\end{figure}
Table~\ref{Table:results} presents the performance comparison results of various methods in determining the appropriate response target in an MHRI scenario. The results indicate that the proposed method (PM) outperforms existing approaches in terms of both accuracy and efficiency. Notably, PM demonstrates a significant improvement in cases where the speaker’s gaze and speaking target are inconsistent (AC (G $\neq$ L): 60.0\%), several times better all other methods. 
Additionally, models such as GPT-4 and GPT-o1-mini, particularly in the text-only settings, exhibit substantially lower performance, highlighting the limitation of relying solely on textual input without multimodal cues. This suggests that integrating video features is crucial for effectively capturing conversational dynamics in MHRI.
We further tested the model's average decision time for each utterance, excluding the data preprocessing stage. In terms of computational efficiency, PM significantly outperforms all other approaches, achieving an average per-decision time of 0.01 s, which is at least an order of magnitude faster than the VLM-based methods. This speed advantage is critical for real-time HRI scenarios, where low-latency responses are necessary for maintaining natural and fluid conversations.
Overall, these results demonstrate the superiority of PM in terms of both accuracy and computational efficiency. By effectively incorporating multimodal features, PM achieves a more robust understanding of conversational context, ensuring that the robot responds appropriately to the intended speaker in real-time interactions.

Table~\ref{Table:ablation} presents the ablation study results, analyzing the impact of multi-task learning and KL divergence losses on model performance. Comparing (a) and (b), training only on ground-truth dialogue data without multi-task learning (a) achieves an average accuracy of 64.5\%, while incorporating $\mathcal{L}^{\text{r}}_{\text{KL}}$ alone (b) does not yield a significant improvement (64.6\%).
Introducing multi-task learning in (c) improves the performance to 65.8\%, demonstrating the benefit of shared representations across tasks. Further adding $\mathcal{L}^{\text{s}}_{\text{KL}}$ in (d) slightly increases the accuracy to 66.0\%, indicating that this KL loss helps refine the model’s decision-making. The best performance is observed when both KL losses are applied (f), achieving an average accuracy of 66.2\%, with a notable improvement in cases where gaze and speaking target are aligned (72.2\%). This suggests that optimizing speaker prediction concerning gaze alignment contributes positively to overall accuracy.

To evaluate the effectiveness of feature representation in our MHRI framework, we visualized the fused features for dialogue recognition and response decision using t-SNE. As shown in Fig.~\ref{Fig:visualize a}, the dialogue recognition features exhibit a relatively dense distribution with partially overlapping clusters, indicating the inherent complexity of distinguishing multiple speakers in a conversation. Conversely, Fig.~\ref{Fig:visualize b} shows a more structured separation in the response decision task, suggesting that the learned features effectively capture the key distinctions necessary for determining when and to whom the robot should respond. This contrast highlights the differing nature of these two tasks; while dialogue recognition requires fine-grained differentiation among speakers, response decision relies on identifying salient cues for engagement.
\section{CONCLUSION AND DISSCUSION}
This paper proposed a Transformer-based multi-task learning framework for MHRI, integrating speaker identification and response decision-making. By leveraging multimodal inputs, including text and visual features, our model effectively captures conversational dynamics. We introduced two novel loss functions to improve scene understanding and response selection, and our approach outperforms heuristic-based and single-task methods, achieving state-of-the-art performance on a newly constructed dataset.

However, since our dataset is newly constructed, its generalizability to real-world interactions is uncertain, and while our framework supports multi-user scenarios, its accuracy in more complex settings requires further validation.
Future work will focus on validating our approach in real-world environments, improving multimodal fusion techniques, and enhancing adaptive response strategies to ensure robustness and scalability in diverse multi-user interactions.
\bibliographystyle{IEEEbib}
\bibliography{refs}
\end{document}